\documentclass[11pt,a4paper]{article}
\usepackage[hyperref]{acl2017}
\usepackage{times}
\usepackage{latexsym}

\usepackage{url}

\aclfinalcopy 
\title{ConceptNet at SemEval-2017 Task 2: Extending Word Embeddings with Multilingual Relational Knowledge}

\author{Robyn Speer \\
  Luminoso Technologies, Inc. \\
  675 Massachusetts Avenue \\
  Cambridge, MA 02139 \\
  {\tt rspeer@luminoso.com} \\\And
  Joanna Lowry-Duda \\
  Luminoso Technologies, Inc. \\
  675 Massachusetts Avenue \\
  Cambridge, MA 02139 \\
  {\tt jlowry-duda@luminoso.com} \\}

\date{}

\begin{document}
\maketitle
\begin{abstract}

    This paper describes Luminoso's participation in SemEval 2017 Task 2,
    ``Multilingual and Cross-lingual Semantic Word Similarity'', with a system
    based on ConceptNet. ConceptNet is an open, multilingual knowledge graph
    that focuses on general knowledge that relates the meanings of words and
    phrases. Our submission to SemEval was an update of previous work that
    builds high-quality, multilingual word embeddings from a combination of
    ConceptNet and distributional semantics. Our system took first place in
    both subtasks. It ranked first in 4 out of 5 of the separate languages, and
    also ranked first in all 10 of the cross-lingual language pairs.

\end{abstract}

\section{Introduction}

ConceptNet 5 \cite{speer2013conceptnet} is a multilingual, domain-general
knowledge graph that connects words and phrases of natural language
(\emph{terms}) with labeled, weighted edges. Compared to other knowledge
graphs, it avoids trying to be a large gazetteer of named entities. It aims most
of all to cover the frequently-used words and phrases of every language, and to
represent generally-known relationships between the meanings of these terms.

The paper describing ConceptNet 5.5 \cite{speer2017conceptnet} showed that it
could be used in combination with sources of distributional semantics,
particularly the word2vec Google News skip-gram embeddings
\cite{mikolov2013word2vec} and GloVe 1.2 \cite{pennington2014glove}, to produce
new embeddings with state-of-the-art performance across many word-relatedness
evaluations. The three data sources are combined using an extension of the
technique known as retrofitting \cite{faruqui2015retrofitting}.  The result is
a system of pre-computed word embeddings we call ``ConceptNet Numberbatch''.

The system we submitted to SemEval-2017 Task 2, ``Multilingual and
Cross-lingual Semantic Word Similarity'', is an update of that system,
coinciding with the release of version 5.5.3 of ConceptNet\footnote{Data and
code are available at \url{http://conceptnet.io}.}.  We added multiple fallback
methods for assigning vectors to out-of-vocabulary words. We also experimented
with, but did not submit, systems that used additional sources of word
embeddings in the five languages of this SemEval task.

This task \cite{semeval2017task2} evaluated systems at their ability to rank
pairs of words by their semantic similarity or relatedness. The words are in
five languages: English, German, Italian, Spanish, and Farsi.  Subtask 1
compares pairs of words within each of the five languages; subtask 2 compares
pairs of words that are in different languages, for each of the ten pairs of
distinct languages.

Our system took first place in both subtasks. Detailed results for our system appear in
Section~\ref{sec:results}.

\section{Implementation}

The way we built our embeddings is based on retrofitting
\cite{faruqui2015retrofitting}, and in particular, the elaboration of it we
call ``expanded retrofitting'' \cite{speer2017conceptnet}. Retrofitting, as
originally described, adjusts the values of existing word embeddings based on a
new objective function that also takes a knowledge graph into account.  Its
output has the same vocabulary as its input. In expanded retrofitting, on the
other hand, terms that are only present in the knowledge graph are added to the
vocabulary and are also assigned vectors.

\subsection{Combining Multiple Sources of Vectors}

As described in the ConceptNet 5.5 paper \cite{speer2017conceptnet}, we apply
expanded retrofitting separately to multiple sources of embeddings (such as
pre-trained word2vec and GloVe), then align the results on a unified vocabulary
and reduce its dimensionality.

First, we make a unified matrix of embeddings, $M_1$, as follows:

\begin{itemize}
    \item Take the subgraph of ConceptNet consisting of nodes whose degree is at least 3. Remove
        edges corresponding to negative relations (such as \emph{NotUsedFor} and \emph{Antonym}).
        Remove phrases with 4 or more words.
    \item Standardize the sources of embeddings by case-folding their terms and $L_1$-normalizing
        their columns.
    \item For each source of embeddings, apply expanded retrofitting over that source with the subgraph
        of ConceptNet. In each case, this provides vectors for a vocabulary of terms that includes
        the ConceptNet vocabulary.
    \item Choose a unified vocabulary (described below), and look up the vectors
        for each term in this vocabulary in the expanded retrofitting outputs. If a vector is missing
        from the vocabulary of a retrofitted output, fill in zeroes for those components.
    \item Concatenate the outputs of expanded retrofitting over this unified vocabulary to give $M_1$.
\end{itemize}

\subsection{Vocabulary Selection}

Expanded retrofitting produces vectors for all the terms in its knowledge graph
and all the terms in the input embeddings. Some terms from outside the
ConceptNet graph have useful embeddings, representing knowledge we would like
to keep, but using all such terms would be noisy and wasteful.

To select the vocabulary of our term vectors, we used a heuristic that takes
advantage of the fact that the pre-computed word2vec and GloVe embeddings we
used have their rows (representing terms) sorted by term frequency.

To find appropriate terms, we take all the terms that appear in the first
500,000 rows of both the word2vec and GloVe inputs, and appear in the first
200,000 rows of at least one of them. We take the union of these with the terms
in the ConceptNet subgraph described above. The resulting vocabulary, of
1,884,688 ConceptNet terms plus 99,869 additional terms, is the vocabulary we
use in the system we submitted and its variants.

\subsection{Dimensionality Reduction}

The concatenated matrix $M_1$ has $k$ columns representing features that may be
redundant with each other. Our next step is to reduce its dimensionality to a
smaller number $k'$, which we set to 300, the dimensionality of the largest
input matrix. Our goal is to learn a projection from $k$ dimensions to $k'$
dimensions that removes the redundancy that comes from concatenating multiple
sources of embeddings.

We sample 5\% of the rows of $M_1$ to get $M_2$, which we will use to find the
projection more efficiently, assuming that its vectors represent approximately
the same distribution as $M_1$.

$M_2$ can be approximated with a truncated SVD: $M_2 \approx U \Sigma^{1/2}
V^T$, where $\Sigma$ is truncated to a $k' \times k'$ diagonal matrix of the $k'$
largest singular values, and $U$ and $V$ are correspondingly truncated to have
only these $k'$ columns.

$U$ is a matrix mapping the same vocabulary to a smaller set of features.
Because $V$ is orthonormal, $U \Sigma$ is a rotation and truncation of the
original data, where each feature contributes the same amount of variance as it
did in the original data.  $U \Sigma^{1/2}$ is a version that removes some of
the variance that came from redundant features, and also is analogous to the
decomposition used by \citet{levy2015embeddings} in their SVD process.

We can solve for the operator that projects $M_2$ into $U \Sigma^{1/2}$:
$$U \Sigma^{1/2} \approx M_2 V \Sigma^{-1/2}$$

$V \Sigma^{-1/2}$ is therefore a $k \times k'$ operator that, when applied on
the right, projects vectors from our larger space of features to our smaller
space of features. It can be applied to any vector in the space of $M_1$, not
just the ones we sampled. $M_3 = M_1 V \Sigma^{-1/2}$ is the projection of the
selected vocabulary into $k'$ dimensions, which is the matrix of term
vectors that we output and evaluate.

\subsection{Don't Take ``OOV'' for an Answer}

Published evaluations of word embeddings can be inconsistent about what to do
with out-of-vocabulary (OOV) words, those words that the system has learned no
representation for. Some evaluators, such as \citet{fasttext}, discard all
pairs containing an OOV word. This makes different systems with different vocabularies
difficult to compare. It enables gaming the evaluation by limiting the
system's vocabulary, and gives no incentive to expand the vocabulary.

This SemEval task took a more objective position: no word pairs may be
discarded. Every system must submit a similarity value for every word pair, and
``OOV'' is no excuse. The organizers recommended using the midpoint of the
similarity scale as a default.

In our previous work with ConceptNet, we eliminated one possible cause of OOV
terms. A term that is outside of the selected vocabulary, perhaps because its
degree in ConceptNet is too low, can still be assigned a vector. When we
encounter a word with no computed vector, we look it up in ConceptNet, find its
neighbors, and take the average of whatever vectors those neighboring terms
have. This approximates the vector the term would have been assigned if it had
participated in retrofitting. If the term has no neighbors with vectors, it
remains OOV.

For this SemEval task, we recognized the importance of minimizing OOV terms,
and implemented two additional fallback strategies for the terms that are still
OOV.

It is unavoidable that training data in non-English languages will be harder
to come by and sparser than data in English. It is also true that some words
in non-English languages are borrowed directly from English, and are therefore
exact cognates for English words.

As such, we used a simple strategy to further increase the coverage of our
non-English vocabularies: if a term is not associated with a vector in matrix
$M_3$, we first look up the vector for the term that is spelled identically in
English. If that vector is present, we use it.

This method is in theory vulnerable to false cognates, such as the German word
{\em Gift} (meaning ``poison''). However, false cognates tend to appear among
common words, not rare ones, so they are unlikely to use this fallback
strategy.  Our German embeddings do contain a vector for ``Gift'', and it is
similar to English ``poison'', not English ``gift''.

As a second fallback strategy, when a term cannot be found in its given
language or in English, we look for terms in the vocabulary that have the given
term as a prefix. If we find none of those, we drop a letter from the end of
the unknown term, and look for that as a prefix. We continue dropping letters
from the end until a result is found. When a prefix yields results, we use the
mean of all the resulting vectors as the word's vector.

\section{Results}

In this task, systems were scored by the harmonic mean of their Pearson and
Spearman correlation with the test set for each language (or language pair in
Subtask 2).  Systems were assigned aggregate scores, averaging their top 4
languages on Subtask 1 and their top 6 pairs on Subtask 2.

\subsection{The Submitted System: ConceptNet + word2vec + GloVe}

The system we submitted applied the retrofitting-and-merging process described
above, with ConceptNet 5.5.3 as the knowledge graph and two well-regarded
sources of English word embeddings. The first source is the word2vec Google
News embeddings\footnote{\url{https://code.google.com/archive/p/word2vec/}},
and the second is the GloVe 1.2 embeddings that were trained on 840 billion
tokens of the Common
Crawl\footnote{\url{http://nlp.stanford.edu/projects/glove/}}.

Because the input embeddings are only in English, the vectors in other
languages depended entirely on propagating these English embeddings via the
multilingual links in ConceptNet.

This system appears in the results as ``Luminoso-run2''. Run 1 was similar, but
it was looking up neighbors in an unreleased version of the ConceptNet graph
with fewer edges from DBPedia in it.

This system's aggregate score on subtask 1 was 0.743. Its combined score on
subtask 2 (averaged over its six best language pairs) was 0.754.

\subsection{Variant A: Adding Polyglot Embeddings}

Instead of relying entirely on English knowledge propagated through ConceptNet,
it seemed reasonable to also include pre-calculated word embeddings in other
languages as inputs. In Variant A, we added inputs from the Polyglot embeddings
\cite{alrfou2013polyglot} in German, Spanish, Italian, and Farsi as four
additional inputs to the retrofitting-and-merging process.

The results of this variant on the trial data were noticeably lower, and when
we evaluate it on the test data in retrospect, its test results are lower as well.
Its aggregate scores are .720 on subtask 1 and .736 on subtask 2.

\subsection{Variant B: Adding Parallel Text from OpenSubtitles}

In Variant B, we calculated our own multilingual distributional embeddings from
word co-occurrences in the OpenSubtitles2016 parallel corpus
\cite{lison2016opensubtitles}, and used this as a third input alongside
word2vec and GloVe.

For each pair of aligned subtitles among the five languages, we combined the
language-tagged words into a single set of $n$ words, then added $1/n$ to the
co-occurrence frequency of each pair of words, yielding a sparse matrix of word
co-occurrences within and across languages. We then used the SVD-of-PPMI
process described by \citet{levy2015embeddings} to convert these sparse
co-occurrences into 300-dimensional vectors.

On the trial data, this variant compared inconclusively to Run 2. We submitted
Run 2 instead of Variant B because Run 2 was simpler and seemed to perform
slightly better on average.

However, when we run variant B on the released test data, we note that it would
have scored better than the system we submitted. Its aggregate scores are .759
on subtask 1 and .767 on subtask 2.

\subsection{Comparison of Results}
\label{sec:results}

\begin{table}[t]
\begin{small}
\begin{tabular}{lrrrrr}
\hline
Eval.       & Base & Ours & $-$OOV & Var. A & Var. B \\
\hline
\em en      & .683 & .789 & .747 & .778 & \bf .796 \\
\em de      & .513 & .700 & .599 & .673 & \bf .722 \\
\em es      & .602 & .743 & .611 & .716 & \bf .761 \\
\em it      & .597 & .741 & .606 & .711 & \bf .756 \\
\em fa      & .412 & .503 & .363 & .506 & \bf .541 \\
\bf Score 1 & .598 & .743 & .641 & .720 & \bf .759 \\
\hline
\em en-de   & .603 & .763 & .696 & .749 & \bf .767 \\
\em en-es   & .636 & .761 & .675 & .752 & \bf .778 \\
\em en-it   & .650 & .776 & .677 & .759 & \bf .786 \\
\em en-fa   & .519 & .598 & .502 & .590 & \bf .634 \\
\em de-es   & .550 & .728 & .620 & .704 & \bf .747 \\
\em de-it   & .565 & .741 & .612 & .722 & \bf .757 \\
\em de-fa   & .464 & .587 & .501 & .586 & \bf .610 \\
\em es-it   & .598 & .753 & .613 & .732 & \bf .765 \\
\em es-fa   & .493 & .627 & .482 & .623 & \bf .646 \\
\em it-fa   & .497 & .604 & .474 & .599 & \bf .635 \\
\bf Score 2 & .598 & .754 & .649 & .736 & \bf .767 \\
\hline
\end{tabular}
\end{small}
\caption{
    Evaluation scores by language. ``Score 1'' and ``Score 2'' are
    the combined subtask scores. ``Base'' is the Nasari baseline,
    ``Ours'' is Luminoso-Run2 as submitted, ``$-$OOV'' removes our OOV
    strategy, and ``Var. A'' and ``Var. B'' are the variants we describe
    in this paper.
}
\label{table:results-table}
\end{table}

The released
results\footnote{\url{http://alt.qcri.org/semeval2017/task2/index.php?id=results}}
show that our system, listed as Luminoso-Run2, got the highest aggregate score
on both subtasks, and the highest score on each test set except the monolingual
Farsi set.

Table~\ref{table:results-table} compares the results per language of the system
we submitted, the same system without our OOV-handling strategies, variants A
and B, and the baseline Nasari \cite{camacho2016nasari} system.

Variant B performed the best in the end, so we will incorporate parallel text
from OpenSubtitles in the next release of the ConceptNet Numberbatch system.

\section{Discussion}

The idea of producing word embeddings from a combination of distributional and
relational knowldedge has been implemented by many others, including
\citet{iacobacci2015sensembed} and various implementations of retrofitting
\cite{faruqui2015retrofitting}. ConceptNet is distinguished by the large
improvement in evaluation scores that occurs when it is used as the source of
relational knowledge. This indicates that ConceptNet's particular blend of
crowd-sourced, gamified, and expert knowledge is providing valuable information
that is not learned from distributional semantics alone.

The results transfer well to other languages, showing ConceptNet's usefulness
as ``multilingual glue'' that can combine knowledge in multiple languages into
a single representation.

Our submitted system relies heavily on inter-language links in ConceptNet that
represent direct translations, as well as exact cognates. We suspect that this
makes it perform particularly well at directly-translated English. It would
have more difficulty determining the similarity of words that lack direct
translations into English that are known or accurate.  This is a weak point of
many current word-similarity evaluations: The words that are vague when
translated, or that have language-specific connotations, tend not to appear.

On a task with harder-to-translate words, we may have to rely more on observing
the distributional semantics of corpus text in each language, as we did in the
unsubmitted variants.

\bibliographystyle{acl_natbib}
\bibliography{conceptnet}

\end{document}